%% file: paper_pathtrack_final.tex
\documentclass[10pt,twocolumn,letterpaper]{article}

\usepackage[table]{xcolor}

\usepackage{iccv}
\usepackage{times}
\usepackage{epsfig}
\usepackage{graphicx}
\usepackage{amsmath}
\usepackage{amssymb}

\usepackage{etex}
\usepackage{graphicx}
\usepackage{amsmath,amssymb} 
\usepackage[utf8]{inputenc}
\usepackage{amsmath}
\usepackage{dsfont}
\usepackage{caption}
\usepackage{multirow}
\usepackage{subcaption}
\usepackage{booktabs,caption}
\usepackage[flushleft]{threeparttable} 
\usepackage{booktabs}
\usepackage[font={small}]{caption}
\usepackage[tableposition=above]{caption}
\usepackage{pifont}
\newlength\figureheight
\newlength\figurewidth

\input{macros.tex}

\newcommand*\CHECK{\ding{51}}

\usepackage[breaklinks=true,bookmarks=false]{hyperref}

\iccvfinalcopy 

\def\iccvPaperID{****} 

\begin{document}

\title{PathTrack: Fast Trajectory Annotation with Path Supervision}

\author{
    \begin{tabular}[t]{c@{\extracolsep{2.0em}}c@{\extracolsep{2.0em}}c@{\extracolsep{2.0em}}c}
      Santiago Manen$^{1}$ & Michael Gygli$^{1}$ & Dengxin Dai$^{1}$ & Luc Van Gool$^{1,2}$
    \end{tabular}
  \\[12pt]
    \begin{tabular}{c@{\extracolsep{1.0em}}c@{\extracolsep{1.0em}}c}
      $^1$Computer Vision Laboratory   &  $^2$ESAT -
      PSI / IBBT\\
      ETH Zurich   & K.U. Leuven  \\
      \multicolumn{2}{c}{ \texttt{\small \{smanenfr,  gygli, daid, vangool\}@vision.ee.ethz.ch} } 
    \end{tabular}
}

\maketitle

\input{abstract}

\vspace*{-5mm}
\section{Introduction}
\seclabel{introduction}
\input{introduction}

\vspace*{-1mm}

\section{Related work}
\seclabel{related}
\input{related}


\vspace*{-1mm}

\section{Trajectory annotation with path supervision}
\seclabel{method}
\input{method}

\vspace*{-1mm}

\section{The PathTrack dataset}
\seclabel{dataset}
\input{dataset}

\vspace*{-1mm}

\section{Experiments}
\seclabel{experiments}
\input{experiments}

\vspace*{-1mm}

\section{Conclusion}
\seclabel{conclusion}
\input{conclusion}

{\small
\bibliographystyle{ieee}
\bibliography{egbib}
}

\end{document}

%% file: macros.tex
\newcommand{\eqlabel}[1]{\label{eq:#1}}
\newcommand{\seclabel}[1]{\label{sec:#1}}
\newcommand{\figlabel}[1]{\label{fig:#1}}
\newcommand{\tablabel}[1]{\label{tab:#1}}

\newcommand{\sect}[2][]{Sec.~\ref{#1sec:#2}}
\newcommand{\fig}[2][]{Fig.~\ref{#1fig:#2}}
\newcommand{\eq}[2][]{Eq.~(\ref{#1eq:#2})}
\newcommand{\tab}[2][]{Tab.~\ref{#1tab:#2}}

\DeclareMathAlphabet{\mathpzc}{T1}{pzc}{m}{n}

\newcommand{\track}{\ensuremath{T}\xspace}
\newcommand{\tracks}{\ensuremath{\mathcal{T}}\xspace}
\newcommand{\trlab}{\ensuremath{y}\xspace}
\newcommand{\point}[2]{\ensuremath{\textbf{p}_{\textbf{#1}}(#2)\xspace}}
\newcommand{\points}{\ensuremath{\mathcal{P}}\xspace}
\newcommand{\detection}{\ensuremath{d}\xspace}
\newcommand{\detections}{\ensuremath{\mathcal{D}}\xspace}
\newcommand{\labels}{\ensuremath{\mathcal{Y}}\xspace}

\newcommand{\timeframe}{\ensuremath{t}\xspace}
\newcommand{\unary}{\ensuremath{U}\xspace}
\newcommand{\pairwise}{\ensuremath{W}\xspace}
\newcommand{\edges}{\ensuremath{\mathcal{E}}\xspace}

\newcommand{\linkdet}{\ensuremath{x}\xspace}
\newcommand{\confcost}{\ensuremath{C}\xspace}
\newcommand{\detscore}{\ensuremath{s}\xspace}
\newcommand{\detcluster}[1]{\ensuremath{G_{#1}\xspace}}

\newlength{\halfwidth}
\newlength{\fullwidth}


%% file: abstract.tex
\begin{abstract}
Progress in Multiple Object Tracking (MOT) has been historically limited by the size of the available datasets.
We present an efficient framework to annotate trajectories and use it to produce a MOT dataset of unprecedented size.
In our novel \emph{path supervision} the annotator loosely follows the object with the cursor while watching the video, providing a \emph{path annotation} for each object in the sequence.
Our approach is able to turn such weak annotations into dense box trajectories.
Our experiments on existing datasets prove that our framework produces more accurate annotations than the state of the art, in a fraction of the time.
We further validate our approach by crowdsourcing the \emph{PathTrack} dataset, with more than 15,000 person trajectories in 720 sequences\footnote{We will provide our dataset and deep models at \url{http://www.project-website.com}.}.
Tracking approaches can benefit training on such large-scale datasets, as did object recognition.
We prove this by re-training an \emph{off-the-shelf} person matching network, originally trained on the MOT15 dataset, almost halving the misclassification rate.
Additionally, training on our data consistently improves tracking results, both on our dataset and on MOT15.
On the latter, we improve the top-performing tracker (NOMT) dropping the number of ID Switches by $18\%$ and fragments by $5\%$.
\end{abstract}

%% file: introduction.tex
\begin{table*}
  \centering
  \caption{Comparison of PathTrack with other popular MOT datasets.}\tablabel{soa_dataset}
  \vspace{-2mm}
  \resizebox{\textwidth}{!}{
  \begin{threeparttable}
  \rowcolors{2}{}{gray!10}
  \begin{tabular}{cccccccccccccc}
  \toprule
  \multirow{2}{*}{Dataset} & \multicolumn{3}{c}{Train} & \multicolumn{3}{c}{Test} & \multicolumn{3}{c}{Total} & \multirow{2}{*}{\begin{tabular}[c]{@{}c@{}}Classes\\ (P = Person, \\ C = Car)\end{tabular}} & \multirow{2}{*}{\begin{tabular}[c]{@{}c@{}}Camera\\ (S=Static\\ M=Moving)\end{tabular}} & \multirow{2}{*}{\begin{tabular}[c]{@{}c@{}}Scene-type\\ label\end{tabular}} & \multirow{2}{*}{\begin{tabular}[c]{@{}c@{}}Camera-\\ movement\\ label\end{tabular}} \\ \cmidrule(lr){2-10}
   & \# seqs & \begin{tabular}[c]{@{}c@{}}Duration\\ (mins)\end{tabular} & \# tracks & \# seqs & \begin{tabular}[c]{@{}c@{}}Duration\\ (mins)\end{tabular} & \# tracks & \# seqs & \begin{tabular}[c]{@{}c@{}}Duration\\ (mins)\end{tabular} & \# tracks &  &  &  &  \\ 
   \midrule

  Virtual KITTI \cite{Gaidon2016} & - & - & - & - & - & - & 5\tnote{*}& 4\tnote{*} & 261\tnote{*} & C\tnote{*} & car-mounted &  &  \\ 

  KITTI \cite{Geiger2012} & 21 & 13 &  & 29 & 18 & - & 50 & 30 & - & C + P & car-mounted &  &  \\ 
  MOT15 \cite{Leal2015} & 11 & 6 & 500 & 11 & 10 & 721 & 22 & 16 & 1221 & P & S+M &  &  \\ 
  MOT16 \cite{Milan2016} & 7 & 4 & 512 & 7 & 4 & 830 & 14 & 8 & 1342 & C+P\tnote{**} & S+M &  &  \\ 
  \textbf{\begin{tabular}[c]{@{}c@{}}PathTrack\\ (ours)\end{tabular}} & \textbf{640} & \textbf{161} & \textbf{15,380} & \textbf{80} & \textbf{11} & \textbf{907} & \textbf{720} & \textbf{172} & \textbf{16,287} & \textbf{P} & \textbf{S+M} & \textbf{\CHECK} & \textbf{\CHECK} \\ 
  \bottomrule
  \end{tabular}
  \begin{tablenotes}
  \item[*] \cite{Gaidon2016} provides 10 different conditions (e.g. different angles, lighting conditions) for each of the 5 sequences. Sequences are virtually rendered.
  \item[**] \cite{Milan2016} provides a rich set of labels, such as whether an object is an occluder or a target is riding a vehicle.
    \vspace{-3mm}
  \end{tablenotes}
  \end{threeparttable}}
  \vspace{-3mm}
\end{table*}

Progress in vision has been fueled by the emergence of datasets of ever-increasing scale.
An example is the surge of Deep Learning thanks to ImageNet \cite{AlexNIPS2012, Olga2015}.
The scaling up of datasets for Multiple Object Tracking (MOT) however has been limited due to the difficulty and cost to annotate complex video scenes with many objects.
As a consequence, MOT datasets consist of only a couple dozens of sequences \cite{Geiger2012,Leal2015,Milan2016} or are restricted to the surveillance scenario \cite{Wen2015}.
This has hindered the development of fully learned MOT systems that can generalize to any scenario.
In this paper, we tackle these issues by introducing a fast and intuitive way to annotate trajectories in videos and use it to create a large-scale MOT dataset.

\begin{figure}[t]
	\scalebox{1}{\includegraphics[width=\linewidth]{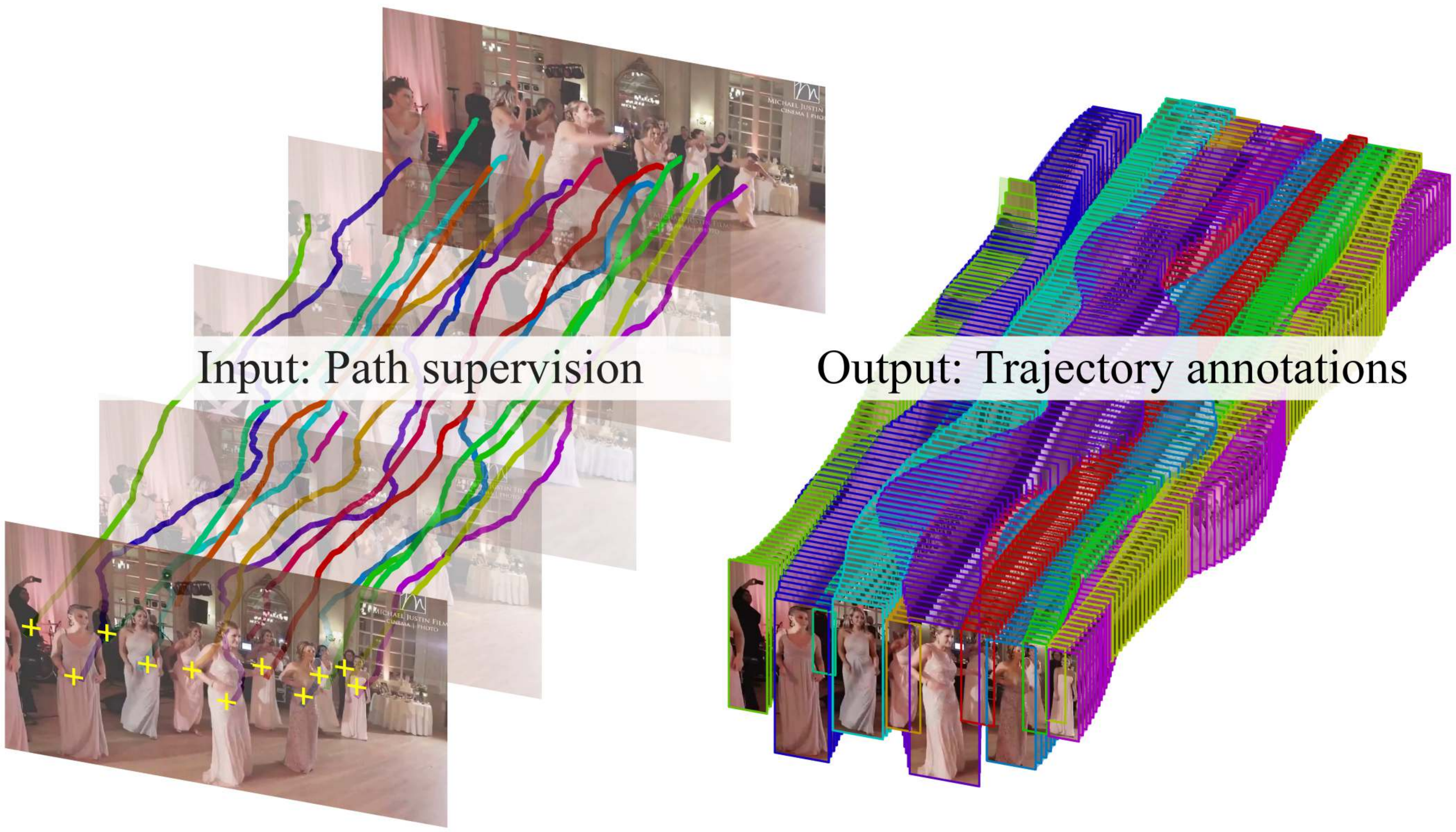}}
    \vspace*{-6mm}
    \caption{This sequence is heavily crowded with similarly-looking people. Annotating such sequences is typically time-consuming and tedious. In our \emph{path supervision}, the user effortlessly follows the object while watching the video, collecting \emph{path annotations}. Our approach produces dense box trajectory annotations from such path annotations.}
\figlabel{fig1}
\vspace*{-6mm}
\end{figure}

Objects can be annotated at different levels of detail.
The cheapest way is to provide video-level object labels \cite{PrestCVPR2012} or action labels \cite{BandlaICCV2013}.
On the other end of the spectrum, sophisticated methods \cite{PerazziCVPR2016, LeeICCV2011, BadrinarayananCVPR2010, VijayanarasimhanECCV2012, FathiBMVC2011} produce pixel-accurate segmentations of objects.
Per-frame bounding box annotations lie in between these extremes.
We call this the \emph{trajectory annotation} task.
The common approach to it is to annotate a sparse set of boxes and interpolate between them linearly \cite{YuenICCV2009} or with shortest-paths \cite{VondrickIJCV2013}.
This is expensive, \eg it cost \emph{tens of thousands of dollars} to annotate the VIRAT dataset \cite{VondrickECCV2010}.

The typical annotation pipeline involves the user idly watching the video in-between manual annotations.
This is arguably a waste of time.
In this paper, we present \emph{path supervision} as a more productive alternative.
In it, the annotator follows the object with the cursor while playing the video, collecting a \emph{path annotation}, \cf~\fig{fig1}.
Hence, \emph{watching time} is efficiently turned into \emph{annotation time}.
Our experiments show that these paths are fast to annotate, almost in real time.

Path annotations are approximate and do not provide the scale of the object.
So recovering full box trajectories from them is far from trivial.
We alleviate these problems by using object detections, since our goal is to generate large MOT datasets, for which we know the class of interest.
Our optimization produces an accurate box trajectory for each path annotation, by linking detections in a global optimization.
Our approach is presently the fastest way to annotate MOT trajectories for any annotation quality.

Since our annotation approach is intuitive, we could crowd source a large-scale dataset with Amazon Mechanical Turk (AMT) \cite{AMT}.
This \emph{PathTrack} dataset is our second major contribution: a large MOT dataset of more than 15,000 person trajectories in 720 sequences, 30 times more than currently available ones \cite{Leal2015}.
Its focus lies on a large-scale and diverse training set, aimed to initiate a new generation of fully data-driven MOT systems.
We show its potential by learning better detection-association models for MOT, which substantially improves the top-performing tracker in MOT15, i.e. NOMT \cite{ChoiICCV2015}.
In summary, our contributions are:
\vspace{-1mm}
\begin{enumerate}
  \itemsep0em
  \item[--] A novel approach to produce full box trajectories from path annotations. It is currently the fastest way to annotate trajectories for any annotation quality and it specially shines for quick quantity-over-quality data collection strategies, ideal for training data.
  \item[--] The novel PathTrack MOT dataset, which includes the collection and annotation of 720 challenging sequences. It focuses on providing abundant training data to learn data-driven trackers. We show its potential by improving the top tracker on MOT15 \cite{Leal2015}.
  \item[--] Insights into collection of training data for MOT. Our experiments show that the MOT community can still benefit from more training data and a saturation point has not yet been reached. Furthermore, quantity seems to be more important than quality when learning to link detections into trajectories.
\end{enumerate}
\vspace{-2mm}

%% file: related.tex
There is quite some work on multimedia annotation \cite{Dasiopoulou2011}.
The most related works annotate objects in videos and can generate datasets for MOT training and evaluation.

\vspace*{-3mm}

\paragraph{Trajectory annotation in videos}

We focus on frameworks aimed at annotating persons with the purpose of generating tracking datasets.
Of less relevance to us are those that work on videos with only a few people, such as \cite{WangICML2014, CastanedaIP2016}.
The naive way to annotate trajectories is to indicate the object location in every frame.
This is inefficient as objects tend to move little between frames.
Hence, VIPER-GT \cite{Mihalcik2003} and LabelMe video \cite{YuenICCV2009} propose to linearly interpolate boxes between annotated keyframes.
There is also a family of methods that learn an appearance model from a sparse set of box annotations. 
VATIC \cite{VondrickECCV2010} uses this appearance model to define a graph on which it performs a shortest-path interpolation between manual annotations with Dynamic Programming \cite{Bellman1954}.
The shortest-path interpolation allows for larger time gaps without manual annotations, assuming that the object is clearly visible, and it can be efficient \cite{WeiICCV2007}.
A VATIC improvement \cite{VondrickNIPS2011} incorporated active learning to decide which frames to annotate, to maximize the gain coming with such frames \cite{Prince2011}.
\cite{CiptadiICCV2015} built on top of shortest-path interpolation by updating the optimization weights with each extra annotation.
Recently, \cite{GilJIVP2016} reconstructed annotated boxes and interpolated the final trajectories in 3D space.
Based on the aforementioned approaches, multiple annotation tools have been developed \cite{Kavasidis2012,BiancoCVIU2015, Schallauer2008}. 
Some gamify the annotation process \cite{DiSalvoACM2013}.
As an alternative to trajectory supervision, some works aim to automatically discover and track objects in video collections, e.g. \cite{KwakICCV2015}.

Compared to previous approaches, we annotate large quantities of videos with the minimum effort possible and prefer quantity over quality in our training data, which have shown success in other tasks.

\vspace*{-3mm}

\paragraph{Path supervision}
Pointing at objects comes very natural and has often been used in human-computer interaction \cite{Hosoya2004,Dutt2016}, yet it only recently gained popularity in Computer Vision.
In parallel with our work, \cite{MettesECCV2016} found path annotations promising for action localization in videos.
Compared to \cite{MettesECCV2016,WeinzaepfelICCV2015}, we annotate dozens of people in highly-crowded sequences, ideal for MOT purposes.
Also recently, \cite{BearmanECCV2016} and \cite{Dutt2016} used point supervision to segment objects in images and videos, resp.
\cite{Dutt2016} uses \emph{multiple} points to segment, by iteratively re-ranking a collection of thousands of object proposals, called \emph{Click Carving}.

We are the first to propose a trajectory annotation framework based on linking detections with path supervision and use it to generate a large MOT dataset.

\vspace*{-3mm}

\paragraph{Tracking datasets}

There is a corpus of video datasets that provide frame-level \cite{caba2015,gygli2015} or pixel-level annotations \cite{PerazziCVPR2016}.
\cite{Kristan2015} and \cite{Real2017} are the largest datasets for single object tracking.
Most large-scale MOT datasets are restricted to surveillance videos \cite{PETS2010, AVSS2011, Wen2015}, since they depict smooth and quasi-linear trajectories that are easy to annotate.
More related to ours, KITTI \cite{Geiger2012} is collected from a car-mounted camera and focuses on pedestrians and vehicles.
Parts of this dataset have been reproduced and rendered virtually, to show the potential of virtual datasets \cite{Gaidon2016}.
\cite{Leal2015,Milan2016} have become the standard benchmarks for MOT, containing complex pedestrian scenes with static or moving cameras.
Compared to these datasets, ours exhibits more diverse scenes and camera movement and is 33 times larger.
\tab{soa_dataset} shows a quantitative comparison.

%% file: method.tex
\begin{figure*}[t]
  \includegraphics[width=\textwidth]{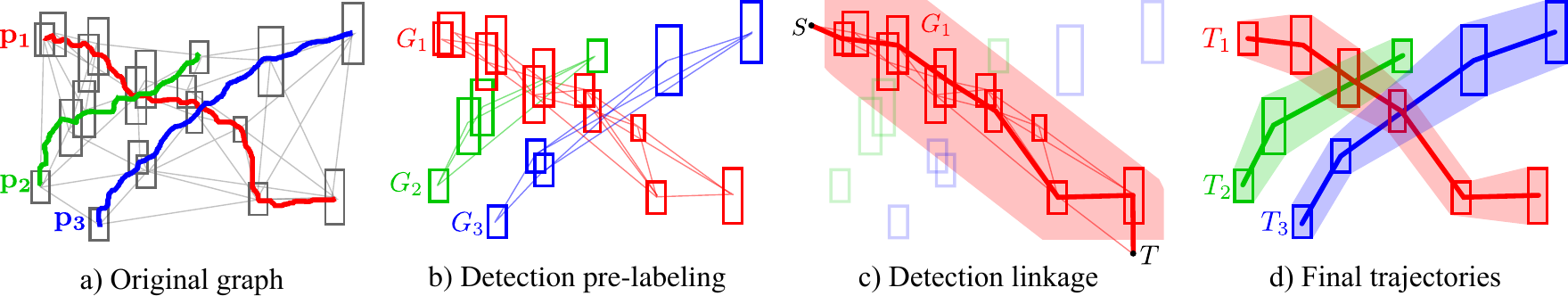}
  \vspace*{-6mm}
  \caption{Overview of our pipeline. a) We take path annotations ($\mathbf{p_1}, \mathbf{p_2}, \mathbf{p_3}$) and object detections as input. b) We pre-label each detection with a potential path candidate, creating detection clusters (\detcluster{1}, \detcluster{2}, \detcluster{3}). c) For each cluster, we compute the most likely trajectory via ST shortest paths. Finally, we output full bounding-box trajectories ($\track_1, \track_2, \track_3$) for each path annotation.}\figlabel{method:overview}
  \vspace*{-6mm}
\end{figure*}

In this section, we describe our annotation framework: we formalize path supervision in \sect{path_supervision} and then detail how we leverage it to infer accurate trajectories in \sect{trajectory_annotation}. In \sect{boxsupervision} we show how to incorporate box supervision.

\subsection{Path supervision}
\seclabel{path_supervision}

A path annotation of an object $i$ consists of an (x, y)-coordinate $\point{i}{\timeframe}$ that \emph{lies} inside its bounding box at frame id $\timeframe$. 
Path annotations are intuitive and efficient to obtain by watching each object independently while following its location with the mouse cursor, \cf~\fig{fig1}.
Our results show that annotating paths is only $33\%$ slower than watching the video in real time.
We say that a video has path supervision if a human annotator has provided a \emph{path annotation} for the objects of interest.
The following section explains how we use these annotations to obtain accurate box trajectories.

\subsection{From path supervision to full box trajectories}
\seclabel{trajectory_annotation}

While path supervision is intuitive and efficient, it comes with its own set of challenges:
a) It offers no information about the spatial extent of the object.
b) The relative position of the path annotation inside the object is unknown.
We partially solve these two problems by drawing on the success of \emph{object detection}, since our final goal is to generate large MOT datasets and we know what kind of objects we want to annotate.
Object detection is gaining maturity for objects of primary interest, so it is natural to use it as an established technique.
Each detection is represented with a box and a confidence score at a given frame.

Our goal is to infer the trajectories \tracks of the objects in the sequence, given the set of input path annotations \points and object detections $\detections$. 
This problem is similar to the tracking-by-detection data association problem, but with additional information from path supervision: the number of objects, their time span and their rough location are given.
Our optimization considers the following intuitive forces:
\begin{enumerate}
\item \textbf{Path potential:} Detections should be assigned to trajectories with compatible path annotations.
\item \textbf{Video-content potential:} \emph{Confident detections} should be used and \emph{affine detections} should be encouraged to have the same label. 
We say that two detections are affine if they are likely to belong to the same object in different frames, according to the content of the video.
\item \textbf{Trajectory constraint:} Trajectories have a single location per frame. Therefore, at most one detection can be assigned to one trajectory at any given frame.
\end{enumerate}
We include these conditions in a two-step optimization.
We first relax the trajectory constraint and label each detection with a provisional trajectory.
This clusters the detections according to their corresponding trajectory, \cf~\fig{method:overview}b.
We can assume that a final trajectory can be constructed with detections from its cluster, and will not contain detections from another cluster.
This \emph{detection pre-labeling step} is detailed in \sect{prelabeling}.
At this point each cluster is might include false positives, which violate the trajectory constraint.
So, in a second step, we find the most probable trajectory in each cluster in a \emph{detection linkage} step, see \fig{method:overview}c.
We describe this step in \sect{linkage}.

\vspace*{-3mm}

\subsubsection{Detection pre-labeling}
\seclabel{prelabeling}

The goal of this step is to assign a path annotation label $\trlab_i$ to each detection $\detection_i$.
Dropping the trajectory constraint allows us encode the path and video-content potentials in a global discrete energy minimization framework.
Intuitively, we will assign path annotations to compatible object detections, assigning affine detections to the same cluster.
The optimal label assignment $\labels^{*}$ is that which minimizes:

\vspace*{-4mm}

\begin{equation}
\begin{aligned}
& \underset{\labels}{\text{minimize}}
& & \sum_{i \in \detections} \unary(\trlab_i) + \sum_{(i, j) \in \edges} \pairwise(\trlab_i, \trlab_j)
\end{aligned}
\eqlabel{labeling}
\vspace*{-3mm}
\end{equation}
where the unary potential $\unary(\trlab_i)$ is the cost of assigning label $\trlab_i$ to detection $i$ and the pairwise potential $\pairwise(\trlab_{i}, \trlab_{j})$ the cost of assigning different labels to detections $i$ and $j$ according to their affinity.
For computational reasons, we limited to a temporal window of 4 seconds, which did not worsen the empirical results.
\fig{method:overview}b illustrates a typical pre-labeling result.
We now describe the potentials we use.

As aforementioned, we do not assume the path annotations to be pixel-accurate center annotations.
Instead we assume that they frequently lie in the bounding box of the object, a much weaker restriction.
Therefore, our unary potential encourages assigning a label $\trlab$ to a detection $\detection_i$ if the corresponding path annotation $\point{\trlab}{t_{i}}$ falls inside the detection for the corresponding frame $t_i$:
\vspace*{-1mm}
\begin{equation}
  \unary(\trlab_i) = \begin{cases}
    0, & \text{if } \point{\trlab}{\timeframe_i} \in \detection_i, \\
    \infty, & \text{otherwise}.
  \end{cases}
  \vspace*{-2mm}
\end{equation}
Indeed our unary only requires a rough location of the path annotation somewhere inside the bounding box of the object.
Note that this requirement does not need to be satisfied in every frame: the path supervision occasionally falling inside the object is usually enough to annotate it accurately.
We prune detections which do not contain path annotations.

While the unary potential is based on the path supervision, the pairwise encodes video content.
It discourages affine detections being assigned to different clusters:
\vspace*{-1mm}
\begin{equation}
  \pairwise(\trlab_i, \trlab_j) = \begin{cases}
    -\log{a_{ij}}, & \text{if } \trlab_i \neq \trlab_j, \\
    0, & \text{otherwise}.
  \end{cases}
  \vspace*{-1mm}
\end{equation}
where $a_{ij}$ represents the affinity between detections $i$ and $j$ and must be decimal number between 0 and 1.
This pairwise potential is submodular, so the energy function \eq{labeling} can be solved with Graph Cuts \cite{Kolmogorov2002} efficiently.
We now describe the affinity measure we used.

\begin{figure*}
    \includegraphics[width=\textwidth]{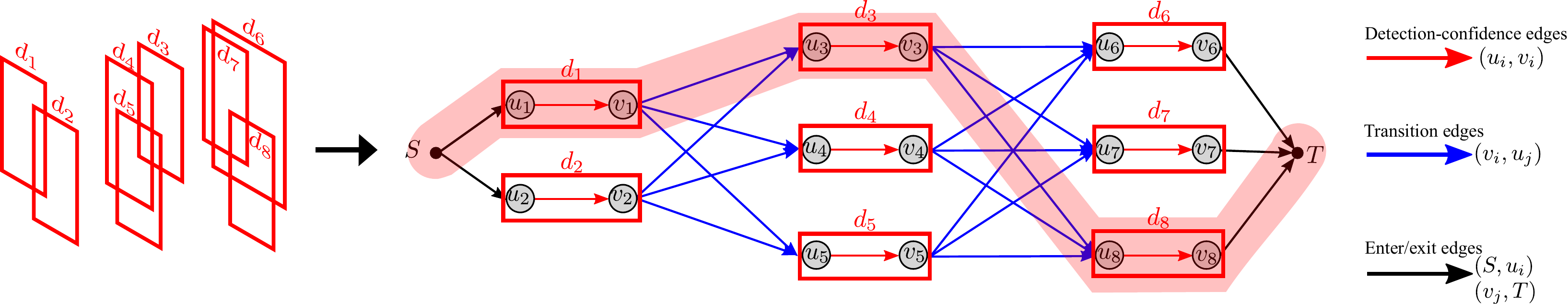}
    \vspace*{-6mm}
    \caption{Over a set of prelabeled detections a min-cost flow network is defined. Each detection is represented by an \emph{observation edge} with the confidence cost. ST-shortest paths are computed over this graph, red shadow.}\figlabel{affinity_spaths}
    \vspace*{-4mm}
\end{figure*}

\vspace*{-3mm}

\paragraph{OF-trajectory affinity measure}
In our work, we use an affinity measure based on \emph{optical-flow trajectories} (OF trajectory).
These are obtained by linking pixels through time using frame-to-frame optical flow and forward-backward consistency checks \cite{Fragkiadaki2012}.
These trajectories are represented with an (x, y)-position for each frame in their time span.
Intuitively, two detections that share many OF trajectories are very likely to belong to the same object.
Thus, we define the affinity between two detections as the intersection-over-union of their OF-trajectories, in the spirit of \cite{ChoiICCV2015}.
More details follow in the supplementary material.

So far we have discussed how we pre-assign object detections to path annotations \cf~\fig{method:overview}b.
In the following section, we describe how to obtain the most likely trajectory for each detection cluster via shortest-paths, \cf~\fig{method:overview}c.

\vspace*{-3mm}

\subsubsection{Detection linkage}
\seclabel{linkage}

In this second step, the goal is to infer the final object trajectories.
Finding the most probable detection-paths in a set of detections has been well studied in the MOT literature \cite{Luo2014}.
We assume that the detection pre-labeling step has labeled the set of detections appropriately.
So each detection can either be part of its assigned trajectory or a false positive, but it can not belong to another trajectory.
Thus, we process each detection-cluster independently \fig{method:overview}c and find the most probable detection-path in the cluster \fig{method:overview}d.

Let $\track_i$ be the final trajectory corresponding to detection-cluster $i$.
It will be composed of a set of time-sorted detections $\linkdet_1$ to $\linkdet_K$.
We find the most likely trajectory by minimizing the sum of detection-confidence costs and between-detections transition costs \cite{ZhangCVPR2008}.
\fig{affinity_spaths} shows how this can be intuitively interpreted as finding the shortest-path in a directed ST-graph where detections are represented by detection-confidence edges.
The optimal detection-path will have the lowest cost:

\begin{equation}
	\vspace{-2mm}
	\underset{\track}{\text{minimize}} \sum_{i = 1}^K \confcost(\linkdet_i) + \sum_{i = 1}^{K - 1} \pairwise(\linkdet_i, \linkdet_{i + 1}) 
    \eqlabel{shortestpaths}
\end{equation}
where $\confcost$ is the detection-confidence cost.
$\confcost_i$ follows the expression $\log((1 - \detscore_i) / \detscore_i)$, where $\detscore_i$ is the 0-to-1 score-confidence of the detection.
Importantly, we use the same transition costs $\pairwise$ when linking detections as we used in step one \eq{labeling} for pre-labeling detections.
Reusing pairwise costs makes the method more efficient.
The detection-confidence costs become negative for confident detections, encouraging the optimization to include them in the final position, while the transition costs penalize the association of detections which are unlikely to be connected.
We refer the reader to \cite{ZhangCVPR2008} for details.
The entry and exit nodes, S and T, are connected only to the earliest and latest detections in the cluster, respectively, ensuring that the trajectory has the same time span as its corresponding path annotation.

As result of the optimization we have a sparse detection-path.
Empirically we find the gap between detections to be small, 0.2 on average in our data. 
Thus we opt to linearly interpolate between detections to obtain the final trajectory, as per standard practice \cite{YuenICCV2009}.

Until now we have presented our annotation approach using path supervision.
It is useful for quickly annotating many sequences, particularly interesting for training data collection.
We propose next an extension to incorporate additional box annotations, improving trajectories up to ground-truth quality.

\subsection{Incorporating box supervision}
\seclabel{boxsupervision}

We propose a simple yet effective way to extend our method with box annotations, to achieve ground-truth quality.
Consider the detection-path we used to interpolate a trajectory.
To include a box annotation, we simply add it to the path and then remove temporally close detections, those less than half a second away.
Interpolating the updated detection-path produces the final trajectory.
These fast updates progressively improve trajectory annotations as more box annotations are included.
Our method is more accurate than the state of the art for any number of updates, as we show in the experiments.

%% file: dataset.tex
We use our annotation approach to collect a MOT dataset of unprecedented size.
This \emph{PathTrack} dataset is an important part of our contribution.
We first provide an overview of the dataset in \sect{dataset_overview}.
Then, we describe how we crowdsourced the annotations in \sect{crowdsourcing}.
We generate the final trajectory annotations with our approach, which associates R-CNN detections \cite{Ren2015} with the help of path supervision.
Importantly, we focus on training data in order to encourage research in fully data-driven trackers.

\subsection{Dataset overview}
\seclabel{dataset_overview}

\begin{figure*}
	\scalebox{1}{\includegraphics[width=\linewidth]{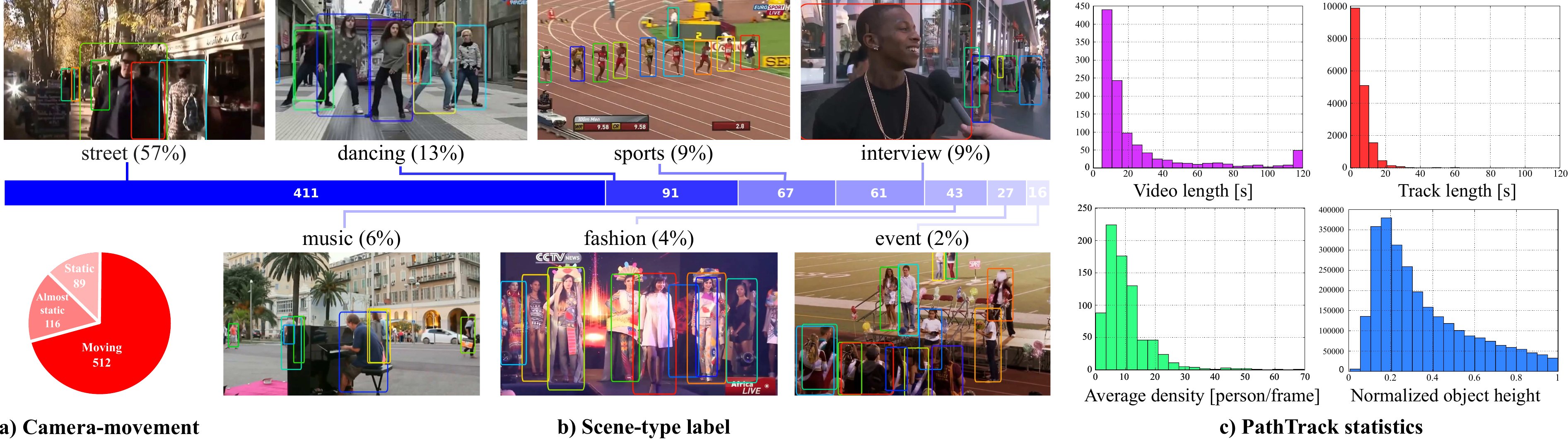}}
    \vspace*{-6mm}
    \caption{Scene-label distribution in PathTrack. We show in a) the distribution of camera-movement labels. Almost three quarters of the sequences have been recorded with a moving camera. We show in b) the distribution of scene-type labels and corresponding examples. More than half of the sequences are street scenes. c) Statistics of PathTrack. Videos are up to 2 minutes long.}\figlabel{dataset_examples}
    \vspace{-5mm}
\end{figure*}

The PathTrack dataset consists of 720 sequences with a total of 16,287 trajectories of \emph{humans}.
Focusing on tracking humans allows us to collect more data for this specific class, which is of great interest both in the MOT community and in practical applications.
The sequences are partitioned in a training set of 640 sequences with 15,380 trajectories and a test set of 80 sequences with 907 trajectories.
Importantly, we allow a certain amount of \emph{noise} in the training set annotations.
This noise stems from inaccuracies in the path supervision and full-trajectory inference and has allowed us to annotate more sequences for a given time budget.
Our experiments show that we can learn strong appearance models from large quantities of data even if the annotations are not perfectly clean (\sect{experiments}).
Indeed, favoring quantity over quality when collecting training data has also been found to be beneficial for other tasks \cite{Xiao2015,gygli2015}. 
Additional effort has been made for test annotations to be clean for evaluation purposes.
\tab{soa_dataset} compares PathTrack with other popular MOT datasets.
Compared to MOT15 \cite{Leal2015}, our dataset contains 33 times more sequences and 26 times more trajectory annotations available.
We hope that the large scale of PathTrack encourages research in more data-driven tracking algorithms.

\vspace*{-3mm}

\paragraph{Dataset diversity}
MOT datasets typically focus on surveillance \cite{Wen2015}, street-scenes \cite{Leal2015,Milan2016} or car-mounted cameras \cite{Geiger2012,Gaidon2016}.
With PathTrack, we aim to explore tracking in new types of sequences.
We have thus collected a diverse set of sequences and we have labeled each one according to two criteria: a \emph{camera-movement label} and one out of 7 \emph{scene labels}, \cf~\fig{dataset_examples}.
There is a clear emphasis in street scenes and moving cameras, due to their challenge, ubiquity and general interest.
But our dataset also allows focusing on static cameras or sequences with a lot of motion, such as \emph{sports} and \emph{dancing}.
These fine-grained categories can also help to evaluate tracking under different conditions.
Additional statistics that show the diversity of our data are presented in \fig{dataset_examples}c.
In the following sections, we describe how we crowdsourced the path annotations and detail in the supplementary material how we collected the videos.

\subsection{Crowdsourcing path annotations}
\seclabel{crowdsourcing}

A critical aspect of any annotation framework is whether it is \emph{easy to use}.
This is an often-overlooked factor that is vital if we want to crowdsource annotations.
Path annotation is intuitive and straightforward.
This has allowed us to crowdsource 16,287 path annotations of PathTrack using AMT.
We now describe our interface and the measures we took to ensure the quality of our annotations.

\vspace*{-3mm}

\paragraph{Interface}

Our interface features a video player with browsing capabilities and a list of the current annotations.
The key difference with other interfaces is that ours records the path of the object by following the cursor.
Additionally, the user easily can speed up and slow down the video, according to the speed of the object.
In our measurements, path annotation was only $30\%$ slower than watching the video in real time.
To further improve our final trajectories, we also asked the workers to provide a bounding box for the first and last appearance of each object and a third one in between.
Since some sequences are very long and can contain dozens of people, the workers were allowed to partially annotate sequences.
This also means that some workers received partially annotated videos and had to continue annotating them. 
This was not much of a challenge with our annotation framework.
We have received very encouraging feedback from our workers, validating the ease of use of our interface and suggesting a potential for gamification. Here are some examples:

\begin{itemize}
\vspace{-2mm}
\item[] \emph{``System was very easy to use and the normal speed was perfect for tracking each subject.''}
\vspace{-2mm}
\item[] \emph{``I really enjoyed your hit. I like to do a lot of annotation work on mechanical turk and thought your interface was, once I got used to it, one of the best I have worked with.''}
\vspace{-2mm}
\vspace{-2mm}
\end{itemize}

\vspace*{-3mm}

\paragraph{Qualification process}
After a short training video, \cf~supplementary material, each worker was asked to qualify by annotating the \emph{TUD-Stadtmitte} sequence.
The qualification certificate was only provided if the path and box annotations were similar to the ground truth up to a certain threshold.
This was checked automatically.

\vspace*{-3mm}

\paragraph{Reviewing process}
If the users are not trained properly or the interface is cumbersome to use, crowdsourced annotations can be erroneous \cite{VondrickECCV2010}.
So we have made an extensive effort to review \textit{every single video} and remove bad annotations.
Videos with missing annotations were sent back to the annotation pool.
We revoked the qualification of workers who continuously provided faulty annotations.
Interestingly, only 3 out of our 81 workers were revoked, while previous work had difficulties collecting annotations of sufficient quality \cite{VondrickIJCV2013}.
This further confirms that path annotation is an engaging and natural way to annotate trajectories.

%% file: experiments.tex
We present our experiments in three parts.
First we evaluate our annotation framework in \sect{annotefficiency}.
We then demonstrate in \sect{personmatching} its impact on training data collection for matching detections, which is a key problem of MOT \cite{ChoiICCV2015} that is shared by most trackers.
We finalize by evaluating the impact of our data on the Multi Object Tracking task.

\subsection{Trajectory annotation efficiency}
\seclabel{annotefficiency}

In this section we evaluate the effectiveness and efficiency of path supervision and compare it to other trajectory annotation approaches.

\vspace*{-3mm}

\paragraph{Dataset description}
We evaluate our method on the MOT15 dataset \cite{Leal2015} since it is most similar to our final goal, the generation of a massive MOT dataset.
This dataset consists of 22 sequences, 11 of which belong to the training set.
The sequences are challenging.
Pedestrians are frequently occluded and some sequences have been recorded with a moving camera.
We evaluate on the 521 trajectories of the training set, for which the ground truth is provided.

\vspace*{-3mm}

\paragraph{State of the art}
We compare to other existing trajectory annotation approaches. LabelMe \cite{YuenICCV2009} is an effective framework based on linear interpolation between box annotations. 
The more sophisticated VATIC \cite{VondrickECCV2010} learns an appearance model from the box annotation, which it uses for a shortest-paths interpolation.
An additional extension of VATIC uses active learning to propose to the user which frame to annotate \cite{VondrickNIPS2011}.

\vspace*{-3mm}

\paragraph{Effectiveness of path supervision}

\begin{figure}
	\scalebox{1}{\includegraphics[width=\linewidth]{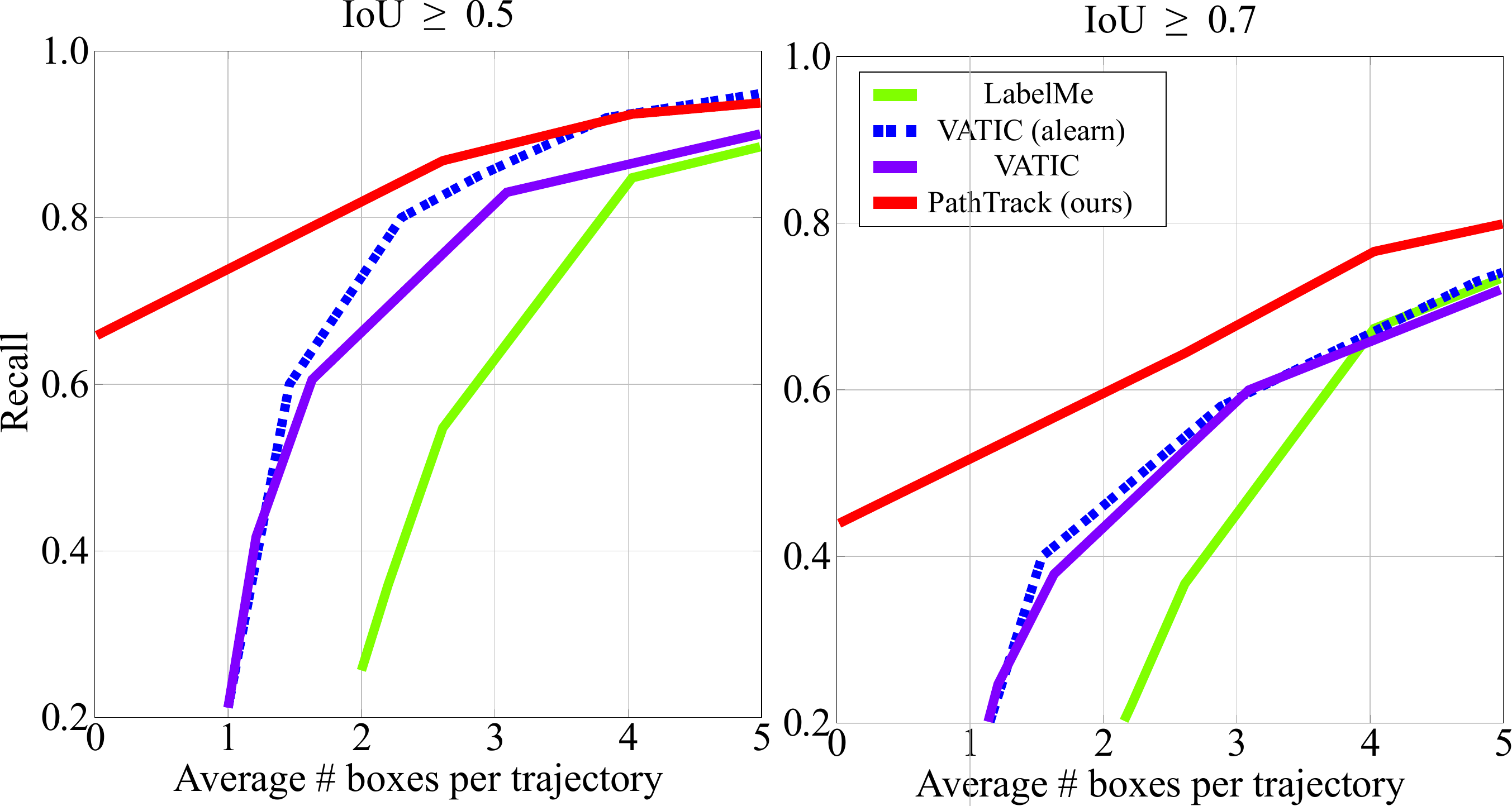}}
    \vspace{-6mm}
    \caption{Performance comparison of 3 state-of-the-art trajectory annotation frameworks and ours. We plot the annotation accuracy for different box-annotation budgets.}\figlabel{acc_nannots}
    \vspace{-4mm}
\end{figure}

We first follow the standard evaluation of trajectory annotation frameworks \cite{VondrickIJCV2013}.
In \fig{acc_nannots}, we compare the annotation accuracy for different amounts of box annotations.
Except for the active learning version of VATIC \cite{VondrickNIPS2011}, box annotations are distributed uniformly in time, \eg, every 10, 5, 1 seconds.
The performance of each framework is measured in terms of how many ground truth boxes are recalled, for different Intersection-over-Union (IOU) \cite{Everingham10} thresholds.
\fig{acc_nannots} demonstrates the \emph{effectiveness} of our path supervision: our cheap path supervision improves performance for any amount of box annotations.
Interestingly, the annotation frameworks seem to converge in performance for large annotation budgets.
A problem of this classical comparison is that it does not take into account the effort required to annotate path trajectories, \ie, it assumes that path annotations can be produced in real time, which is not always the case.
We address evaluate time performance in the next section.

\vspace*{-3mm}

\paragraph{Annotation efficiency}

\begin{figure}
	\scalebox{1}{\includegraphics[width=\linewidth]{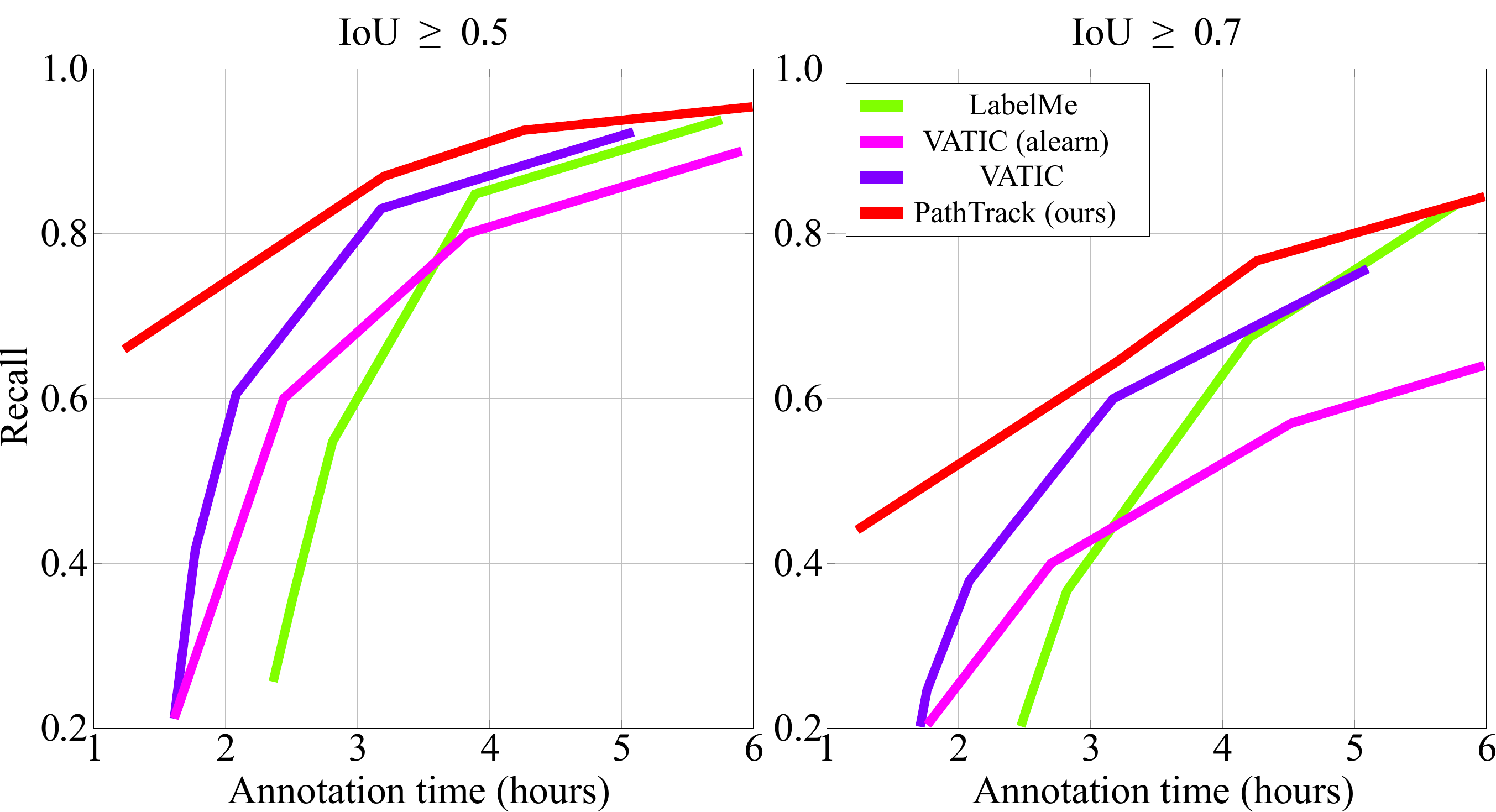}}
    \vspace{-6mm}
    \caption{We compare the efficiency of our method with the state of the art for $0.5$ and $0.7$ IoU thresholds. The time measurements derive from a user study with 91 subjects.}\figlabel{acc_time}
    \vspace{-5mm}
\end{figure}

We compare the \emph{efficiency} of path supervision with previous approaches using a common unit to measure effort: the \emph{annotation time}.
Our time measurements are based on a user study of 78 AMT workers and 13 vision-expert annotators.
We consider three time-consuming components: 1) watching the video at least once to identify the objects, 2) following each trajectory individually while annotating its boxes or path (for ours) and 3) the time required to annotate the bounding boxes.
Our measurements revealed that box annotations take $5.2$ seconds on average and that path annotations require slowing down the video by $33\%$ on average.
We provide a detailed explanation in the supplementary material.
We use these time measurements to produce \fig{acc_time}, where we compare the efficiency of our framework with the state of the art.
Our method is efficient, as VATIC \cite{VondrickECCV2010} and LabelMe \cite{YuenICCV2009} respectively require almost twice and three times more time to obtain our accuracy with only path supervision.
We observe again how all methods converge to the same performance for a larger annotation budget, but ours is much more accurate for very small annotation-budgets.

Overall, our framework is ideal for fast video annotation, which is desirable for generating large training sets, as we demonstrate in the next section.

\begin{figure*}
	\scalebox{1}{\includegraphics[width=\linewidth]{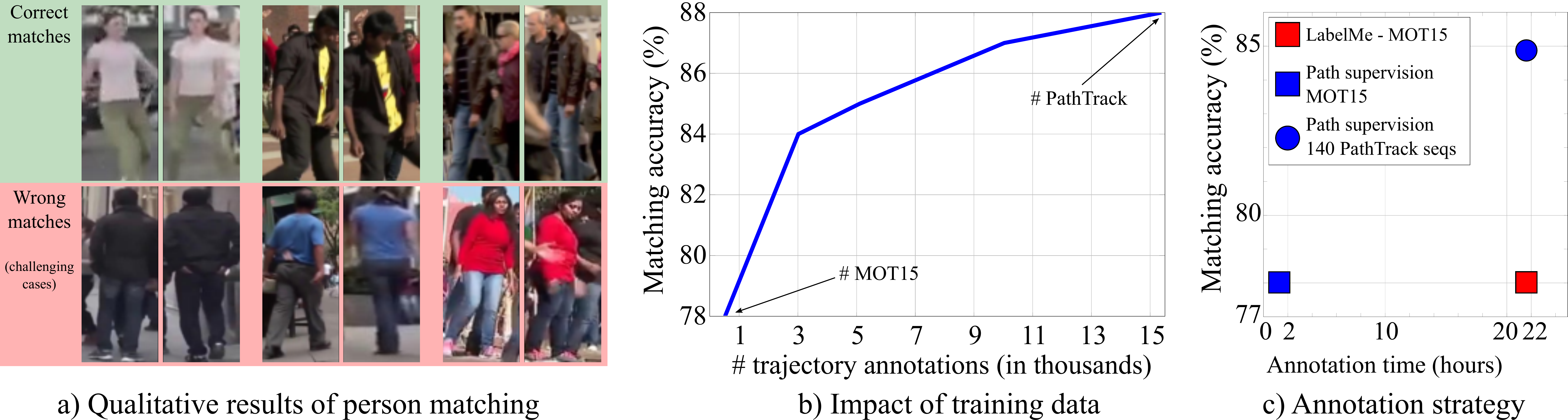}}
    \vspace{-6mm}
    \caption{We show in a) qualitative results of our person matcher on PathTrack. False positives are even challenging for humans. b) Evolution of matching accuracy for different amounts of training trajectories. Training on the 15,380 trajectories of PathTracks results in an accuracy of 88\%, reducing the misclassification rate by 45\%, compared to MOT15. c) Person-matching accuracy for different annotation times using path supervision (\textcolor{blue}{blue}) or exhaustive LabelMe annotation (\textcolor{red}{red}). A high-quantity annotation strategy with our path supervision provides the best accuracy for the same annotation-time budget.}\figlabel{qual_quant}
    \vspace{-5mm}
\end{figure*}

\subsection{Person matching}
\seclabel{personmatching}

We demonstrated in the previous section that path supervision is an efficient way to obtain accurate annotations in a short amount of time.
We now explore the implications for a key task in MOT applications: \emph{person matching}.
This key problem consists of determining the likelihood that two detections belong to the same object in different frames \fig{qual_quant}a.
There is a long tradition of handcrafted matching functions in the literature, with Convolutional Neural Networks (CNN) becoming more popular in the last few years.
These models require extensive training data \cite{AlexNIPS2012, Xiao2015}, which we can provide with PathTrack.
Learning tracker-specific components (e.g. entry/exit costs, mixing coefficients) is outside of the scope of this paper, but should be possible with our data.

We aim to answer the questions: 
i) does the tracking community benefit from more training data?, ii) for a limited budget, should we prioritize data quantity or quality? 

\vspace*{-3mm}

\paragraph{Experimental protocol}
We base our conclusions on a person matching network similar to SiameseCNN~\cite{Leal2016}.
The network takes as input the crops of the two detections, resized to 121x53, and outputs a confidence score that they belong to the same object.
These input crops are stacked, so the input volume is of 121x53x6.
The network has a simple AlexNet style architecture of 3 convolutional and 2 fully connected layers~\cite{Leal2016}.
In our evaluations, we sample 2 million training and test samples.
Positives are randomly sampled pairs of detections that belong to the same object up to 6 seconds away.
For each positive we sample a negative pair belonging to another trajectory in the same video.
We use a learning rate of $0.001$.
In our experiments, we train this network with different data sources and compare their test accuracies.
Accuracy refers to the percentage of properly classified pairs.
We evaluate on the test set of PathTrack, for which the ground truth annotation is clean.

\vspace*{-3mm}

\paragraph{Impact of training data}
In \fig{qual_quant}b we evaluate how the accuracy evolves as more training data becomes available.
The left extreme corresponds to training on the 521 trajectories of the MOT15, which yields an accuracy of 78\%.
Training on the full 15,380 trajectories of PathTrack we improve the accuracy by 10\%, \emph{almost halving} the misclassification rate.
This clearly shows the potential of PathTrack.
Moreover, we observe a certain effect of diminishing returns, but have not reached a plateau.
If we use \emph{context features} (e.g. relative distance, size) \cite{Leal2016} in the network, we also see an improvement when using our data, from $84\%$ to $90\%$.
This shows that our data is useful to learn data-driven MOT.

\vspace*{-3mm}

\paragraph{Quantity-over-quality annotation}
When collecting and annotating data for training purposes, a vital question is whether we should coarsely annotate a large amount of data or precisely annotate a small amount of data.
That is, whether we should follow a \emph{quantity} or a \emph{quality} strategy.
We estimate that it would take 22h to perfectly annotate the 11 videos in the training set of the MOT Challenge with LabelMe \cite{YuenICCV2009}.
We reach this number by counting only the number of windows necessary to obtain an accuracy larger than 0.95 IoU.
This represents the \emph{high-quality} strategy.
We compare this with a \emph{high-quantity} strategy, in which, for the same annotation time, we annotate 140 videos of PathTrack with our framework, with path supervision and 3 boxes per trajectory. 
We show the results in \fig{qual_quant}b.
A \emph{high quantity} approach boosts the final accuracy from $78\%$ to $85\%$.
Interestingly, we can also use our method to quickly annotate the MOT 15 training set and train a model with \emph{exactly} the same accuracy \cf~\fig{qual_quant}b.
These results further showcase the benefit of our framework, which is ideal for fast annotation of large datasets.
Other works \cite{Xiao2015,gygli2015} have also found a quantity strategy to be advantageous to train deep models.

\subsection{Multi Object Tracking}
\seclabel{motresults}

In the previous section we demonstrated how we can train strong person-matching models with PathTrack.
We now evaluate what impact this improvement has on MOT performance.
We first use a standard tracker based on Linear Programming (LP) \cite{ZhangCVPR2008} and evaluate it on the test set of PathTrack with the standard CLEAR MOT metrics \cite{Bernardin2008}.
In \tab{motresults}a We compare the performance of this tracker with two different person-matching models: one trained on MOT15 and the other on our data.
Training on PathTrack substantially improves all the metrics.
These also represent the first tracking results on our dataset.
We further show the potential of PathTrack by improving the top-performing tracker in MOT15 \cite{ChoiICCV2015} with our person-matching model \cf~\tab{motresults}b.
More specifically, we use our discriminative person matcher to further link their trajectory results through occlusions, improving the number of ID Switches by $18\%$ with $5\%$ less fragments.
Low-level details about the trackers are presented in the supplementary material.

\begin{table}
  \caption{\small{We show in a) how training on PathTrack improves all metrics compared to training on MOT15. We use in b) our person-matcher (TRID) to improve the top method in MOT15.}}\tablabel{motresults}
  \vspace{-3mm}
  \begin{subtable}{\linewidth}
  \centering
  \caption{Tracking results on PathTrack}
  \vspace{-2mm}
  \resizebox{\columnwidth}{!}{
    \rowcolors{2}{}{gray!10}
    \begin{tabular}{ccccccccc}
    \toprule
    LP Tracker trained on & MOTA $\uparrow$ & MOTP $\uparrow$	& MT $\uparrow$ & ML $\downarrow$ & FP $\downarrow$ & FN $\downarrow$ & ID Switch $\downarrow$ &\\     
    \midrule
    MOT15 \cite{Leal2015}& 24.5 & 81.4 & 44.2\% & 19.2\% & 42,502 & 37,720 & 1,827   \\
    \textbf{PathTrack (ours)} & \textbf{27.6} & \textbf{81.5} & \textbf{47.3\%} & \textbf{18.2\%} & \textbf{40,614} &\textbf{ 36,508} & \textbf{1,576}\\
    \bottomrule
    \end{tabular}
    }
  \end{subtable}
  \begin{subtable}{\linewidth}
  \centering
  \vspace{1mm}
  \caption{Tracking results on MOT15}
  \vspace{-2mm}
  \resizebox{\columnwidth}{!}{
    \rowcolors{2}{}{gray!10}
    \begin{tabular}{ccccccccc}
    \toprule
    Tracker & MOTA $\uparrow$ & MOTP $\uparrow$	& MT $\uparrow$ & ML $\downarrow$ & FP $\downarrow$ & FN $\downarrow$ & ID Switch $\downarrow$ & Frag $\downarrow$ \\     
    \midrule
    NOMTwSDP \cite{ChoiICCV2015}                                              & 55.5 & \textbf{{76.6}} & 39.0\% & \textbf{25.8\%} & \textbf{{5,594}} & 21,322 & 427       & 701   \\
    \begin{tabular}[c]{@{}c@{}}\textbf{+ TRID (ours)}\end{tabular} & \textbf{55.7} & {76.5} & \textbf{40.6\%} & \textbf{{25.8}}\% & {6,273} &  \textbf{{20,611}}  &  \textbf{{351}}  &   \textbf{{667}}\\
    \bottomrule
    \end{tabular}
    }
  \end{subtable}
  \vspace{-4mm}
\end{table}

%% file: conclusion.tex
In this work, we propose a new framework to annotate trajectories in videos using \emph{path supervision}, with the goal of generating massive MOT datasets.
In the path supervision paradigm, the user annotates the position of the objects of interest with the cursor while watching the video.
Our user study shows that this operation is efficient.
Our optimization takes path annotations and object detections and outputs accurate box-trajectories.
We show in our experiments that we can quickly generate large datasets with our path supervision.
We use our approach to annotate PathTrack, a crowdsourced MOT dataset 33 times larger than currently available ones.
Our experiments show that we can improve current person-matching deep models using our data and that this has an impact on MOT accuracy.
We release PathTrack to promote research in richer and more complete tracking models.